\documentclass[10pt,twocolumn,letterpaper]{article}

\usepackage{cvpr}              
\usepackage{multirow} 
\usepackage{makecell}
\usepackage{caption,tabularx,booktabs}
\usepackage[pagebackref,breaklinks,colorlinks,citecolor=cvprblue]{hyperref}
\usepackage{amsmath}
\usepackage{amssymb}
\usepackage{graphicx, fleqn, tabularx, wrapfig}
\usepackage{textcomp}
\usepackage{enumerate}
\usepackage{subcaption}
\usepackage{booktabs}
\usepackage{makecell}
\usepackage{multirow}
\usepackage{nicefrac,xfrac}
\usepackage{listings}
\usepackage{xcolor}
\usepackage{pgfplots}
\usepackage{stackengine}
\usepackage{xspace}
\usepackage{bbm}
\usepackage{bbding}
\usepackage{mathtools}
\usepackage{cancel}
\usepackage{tabularray}
\usepackage{siunitx}
\usepackage[accsupp]{axessibility}

\definecolor{codegreen}{rgb}{0,0.6,0}
\definecolor{codegray}{rgb}{0.5,0.5,0.5}
\definecolor{codepurple}{rgb}{0.58,0,0.82}
\definecolor{backcolour}{rgb}{0.95,0.95,0.92}

\lstdefinestyle{mystyle}{
    commentstyle=\color{codegreen},
    keywordstyle=\color{magenta},
    numberstyle=\tiny\color{codegray},
    stringstyle=\color{codepurple},
    basicstyle=\ttfamily\scriptsize,
    breakatwhitespace=false,
    breaklines=true,
    captionpos=b,
    keepspaces=true,
    numbersep=5pt,
    showspaces=false,
    showstringspaces=false,
    showtabs=false,
    tabsize=2,
    frame=single
}
\makeatletter
\renewcommand{\paragraph}{%
  \@startsection{paragraph}{4}%
  {\z@}{0.4ex \@plus 1ex \@minus .2ex}{-1em}%
  {\normalfont\normalsize\bfseries}%
}
\makeatother

\newcolumntype{L}[1]{>{\raggedright\let\newline\\\arraybackslash\hspace{0pt}}m{#1}}
\newcolumntype{C}[1]{>{\centering\let\newline\\\arraybackslash\hspace{0pt}}m{#1}}
\newcolumntype{R}[1]{>{\raggedleft\let\newline\\\arraybackslash\hspace{0pt}}m{#1}}
\newcolumntype{Y}{>{\centering\arraybackslash}X}

\definecolor{mypurple}{RGB}{223, 185, 226}
\definecolor{myblue}{RGB}{166, 189, 218}

\definecolor{cvprblue}{rgb}{0.21,0.49,0.74}
\usepackage[pagebackref,breaklinks,colorlinks,citecolor=cvprblue]{hyperref}

\def\paperID{65} 
\def\confName{CVPRW\xspace}
\def\confYear{2024\xspace}

\title{MVDiff: Scalable and Flexible Multi-View Diffusion for 3D Object Reconstruction from Single-View}

\author {Emmanuelle Bourigault\\
Visual Geometry Group, Department of Engineering Science, University of Oxford\\
{\tt\small emmanuelle@robots.ox.ac.uk}
\and Pauline Bourigault\\
Department of Computing, Imperial College London\\
{\tt\small p.bourigault22@imperial.ac.uk}
}

\begin{document}
\maketitle
\begin{abstract}
Generating consistent multiple views for 3D reconstruction tasks is still a challenge to existing image-to-3D diffusion models. Generally, incorporating 3D representations into diffusion model decrease the model's speed as well as generalizability and quality. This paper proposes a general framework to generate consistent multi-view images from single image or leveraging scene representation transformer
and view-conditioned diffusion model. In the model, we introduce epipolar geometry constraints and multi-view attention to enforce 3D consistency. From as few as one image input, our model is able to generate 3D meshes surpassing baselines methods in evaluation metrics, including PSNR, SSIM and LPIPS. 
\end{abstract}
    
\section{Introduction}
\label{sec:intro}

Consistent and high-quality novel view synthesis of real-world objects from a single input image is a remaining challenge in computer vision. There is a myriad of applications in virtual reality, augmented reality, robotic navigation, content creation, and filmmaking. Recent advances in the field of deep learning such as diffusion-based models ~\cite{ho2020DDPM,Song2020ScoreBasedGM,song2020improved,text2mesh, blattmann2023align} 
significantly improved mesh generation by denoising process from Gaussian noise. Text-to-image generation has shown great progress with the development of efficient approaches as generative adversarial networks~\cite{brock2018large, karras2021alias, goodfellow2020generative}, autoregressive transformers~\cite{van2017neural, razavi2019generating, esser2021taming}, and more recently, diffusion models~\cite{ramesh2022dalle2, saharia2022imagen, ho2022classifier, ho2022cascaded}. DALL-E 2~\cite{ramesh2022dalle2} and Imagen~\cite{saharia2022imagen} are such models capable of generating of photorealistic images with large-scale diffusion models. Latent diffusion models~\cite{rombach2022StableDiffusion} apply the diffusion process in the latent space, enabling for faster image synthesis. 

Although, image-to-3D generation has shown impressive results, there is still room for improvement in terms of consistency, rendering and efficiency. Generating 3D representations from single view is a difficult task. It requires extensive knowledge of the 3D world. Although diffusion models have achieved impressive performance, they require expensive per-scene optimization.

Zero123~\cite{liu2023zero} proposes a diffusion model conditioned on view features and camera parameters trained on persepective images~\cite{objaverse}. However, the main drawback is the lack of multiview consistency in the generation process impeding high-quality 3D shape reconstruction with good camera control. SyncDreamer~\cite{liu2023syncdreamer} proposes a 3D feature volume into the Zero123~\cite{liu2023zero} backbone to improve the multiview consistency. However, the volume conditioning significantly reduces the speed of generation and it overfits to some viewpoints, with 3D shapes displaying distortions.

In this paper, we present MVDiff, a multiview diffusion model using epipolar geometry and transformers to generate consistent target views. The main idea is to incorporate epipolar geometry constraints in the model via self-attention and multi-view attention in the UNet to learn the geometry correspondence. We first need to define a scene transformation transformer (SRT) to learn an implicit 3D representation given a set of input views. Then, given an input view and its relative camera pose, we use a view-conditioned diffusion model to estimate the conditional distribution of the target view. 

We show that this framework presents dual improvements compared to existing baselines in improving the 3D reconstruction from generated multi-view images and in terms of generalization capability.

In summary, the paper presents a multi-view generation framework from single image that is transferable to various datasets requiring little amount of changes. We show high performance on the GSO dataset for 3D mesh generation. The model is able to extrapolate one view image of a 3D object to 360-view with high fidelity. Despite being trained on one dataset of natural objects, it can create diverse and realistic meshes.
We summarise our contributions as follows: \\
$\bullet$ Implicit 3D representation learning with geometrical guidance \\
$\bullet$  Multi-view self-attention to reinforce view consistency \\ 
$\bullet$ Scalable and flexible framework

\begin{figure*}
  \centering
    \includegraphics[width=1\linewidth]{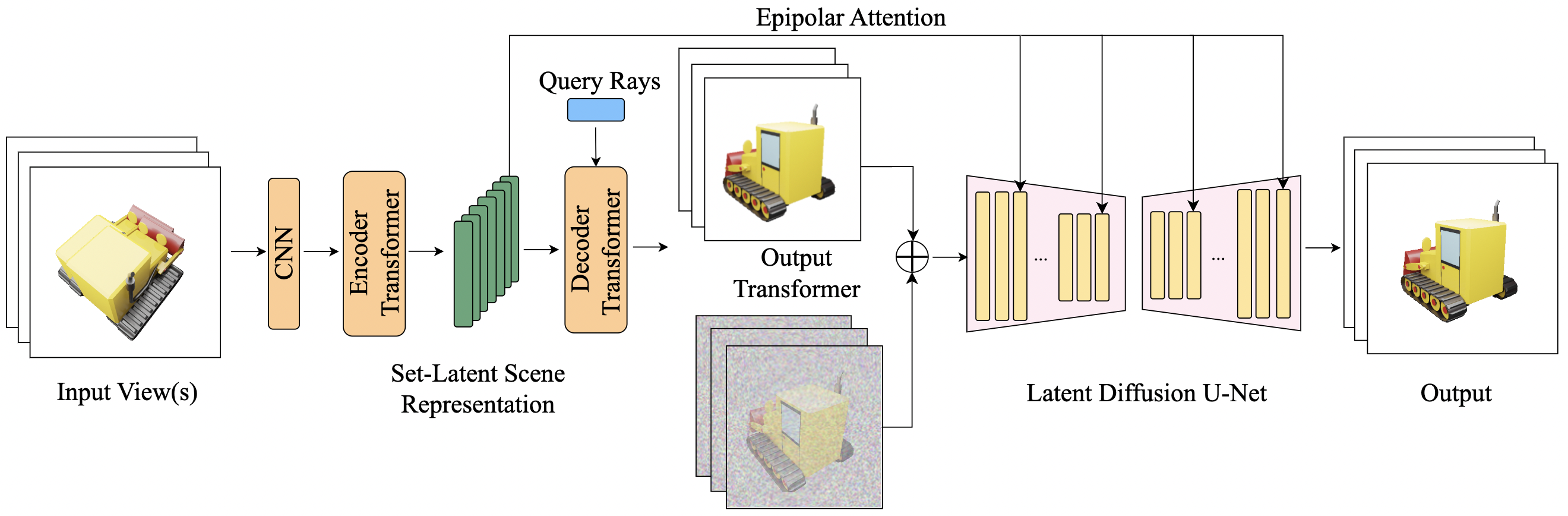}
  \caption{\textbf{Pipeline of MVDiff}. From a single input or few input images, the transformer encoder translates the image(s) into latent scene representations, implicitely capturing 3D information. The intermediate outputs from the scene representation transformer are used as input by the view-conditioned latent diffusion UNet, generating multi-view consistent images from varying viewpoints.}
  \label{fig:Pipeline}
\end{figure*}

\section{Related Work}
\label{sec:related_work}
\subsection{Diffusion for 3D Generation}
Recently, the field of 3D generation has demonstrated rapid progress with the use of diffusion models. Several studies showed remarkable performance by training models from scratch on large datasets to generate point clouds~\cite{Luo2021DiffusionPM,nichol2022pointe},  meshes~\cite{gao2022get3d,liu2023meshdiffusion} or neural radiance fields (NeRFs) at inference. Nevertheless, these models lack generalizability as they are trained on specific categories of natural objects.  DreamFusion~\cite{poole2022dreamfusion} explored leveraging 2D priors to guide 3D generation. Inspired by DreamFusion,  several studies adopted a similar pipeline using distillation of a pretrained 2D text-to-image generation model for generating 3D shapes~\cite{anciukevičius2024renderdiffusion,cao2023largevocabulary,cheng2023sdfusion,müller2023diffrf,wang2022rodin}. The per-scene optimisation process typically lacks in efficiency with times ranging from minutes to hours to generate single scenes.

Recently, 2D diffusion models for multi-view synthesis from single view have raised interest for their fast 3D shape generation with appealing visuals~\cite{liu2023one2345,liu2023zero,shi2023zero123++}. However, they generally do not consider consistency of multi-view in the network design. Zero123 proposes relative viewpoint as conditioning in 2D diffusion models, in order to generate novel views from a single image \cite{liu2023zero}. However, this work does not consider other views in the learning process and this causes inconsistencies for complex shapes. One-2-3-45~\cite{liu2023one2345} decodes signed distance functions (SDF)~\cite{park2019deepsdf} for
3D shape generation given multi-view images from
Zero123 \cite{liu2023zero}, but the 3D reconstruction is not smooth and artifacts are present.

More recently, SyncDreamer~\cite{liu2023syncdreamer} suggests a 3D global feature volume, in order to tackle inconsistencies in multi-view generation. 3D volumes are used with depth-wise attention for maintaining multi-view consistency. The heavy 3D global modeling tend to reduce the speed of the generation and quality of the generated meshes. MVDream~\cite{shi2023mvdream} on the other hand incorporates 3D self-attention with improved generalisability to unseen datasets. EscherNet~\cite{Kong2024EscherNetAG} proposed to leverage camera positional encoding (CaPE) in a transformer-based diffusion model to implicitely learn 3D representations with impressive generalisation and consistency.

\subsection{Sparse-View Reconstruction}
Sparse-view image reconstruction~\cite{jiang2023leap, yang2022fvor} is a challenging task where only a limited number of images, generally less than 10, are given. Traditional 3D reconstruction methods start by estimating camera poses, then as a second step perform dense reconstruction with multi-view stereo~\cite{stereopsis2010accurate, yao2018mvsnet} or NeRF~\cite{NeuS}. Estimating camera poses in the context of sparse-view reconstruction is a challenging task as there is little or no overlap between views. \cite{yang2022fvor} aimed at addressing this challenge by optimising camera poses and 3D shapes simultaneously. In the same line of research, PF-LRM~\cite{wang2023pf-lrm} suggests a pose-free approach to tackle the uncertainty in camera poses.
In our work, we learn the relative camera poses of the 3D representation implicitly via a transformer encoder-decoder network and a view-conditioned diffusion model capable of generating consistent multi-view images directly. We then employ a reconstruction system Neus~\cite{wang2021neus} to recover a mesh.

\section{Methodology}
\label{sec:Methodology}

\subsection{Multi-view Conditional Diffusion Model}

The rationale behind multi-view conditioning in diffusion models is to infer precisely the 3D shape of an object with the constraint that  regions of the 3D object are unobserved. Direct 3D predictions for sequential targets as in Zero123~\cite{liu2023zero} might lead to implausible novel views. To control the uncertainty in novel view synthesis, we choose to enforce multi-view consistency during training.

Given an input image or sparse-view input images of a 3D object, denoted as $\boldsymbol{x}_{\mathrm{I}}$, with known camera parameters $\boldsymbol{\pi}_{\mathrm{I}}$, and target camera parameters $\boldsymbol{\pi}_{\mathrm{T}}$, our aim is to synthesize novel views that recover the geometry of the object.

Our framework can be broken down into two  parts: (i) first a scene representation transformer (SRT)~\cite{sajjadi2022scene} that learns the latent 3D representation given a single or few input views, and (ii) second a view-conditioned diffusion model to generate novel views.

\subsection{Novel View Synthesis via Epipolar Geometry}
\label{subsec:geo-nvs}

To perform novel view synthesis, we employ a scene representation transformer (SRT)~\cite{sajjadi2022scene}. In the work of \cite{sajjadi2022scene}, a transformer encoder-decoder architecture learns an implicit 3D latent representation given a set of images with camera poses $\left(\boldsymbol{x}_{\mathrm{I}}, \boldsymbol{\pi}_{\mathrm{I}}\right)$. First, a CNN extracts features from $\boldsymbol{x}_{\mathrm{I}}$ and feeds them as tokens to the transformer encoder $f_{\textit{E}}$. The transformer encoder then outputs a set-latent scene representation $\boldsymbol{z}$ via self-attention.

For novel view rendering, the decoder transformer of SRT queries the pixel color via cross-attention between the ray associated to that pixel $\boldsymbol{r}$ and the set-latent scene representation $\boldsymbol{z}$.

The aim is to minimize the pixel-level reconstruction loss in \cref{eq:rec_loss},
\begin{align}
  \label{eq:rec_loss}
  \mathcal{L}_{\mathrm{recon}} =\sum_{\mathbf{r} \in \mathcal{R}}\left\|C(\mathbf{r})-\hat{C}(\mathbf{r})\right\|_2^2,
\end{align}
where $C(\mathbf{r})$ is the ground truth color of the ray and $\mathcal{R}$ is the set of rays sampled from target views.

We aim to leverage cross-interaction between images through relative camera poses using epipolar geometrical constraints. For each pixel in a given view $i$, we compute the epipolar line and the epipolar distance for all pixels in view $j$ to build a weighted affinity matrix $A_{i, j}^\prime = A_{i, j} + W_{i, j}$ where $W_{i, j}$ is the weighted map obtained from the inverse epipolar distance.
\paragraph{View-Conditioned Latent Diffusion.} The outputs from SRT do not recover fine details with simple pixel-level reconstruction loss. We employ a view-conditioned diffusion model LDM from \cite{LDM} to estimate the conditional distribution of the target view given the source view and the relative camera pose: $p \left(\boldsymbol{x}_{\mathrm{\textit{T}}} \mid \boldsymbol{\pi}_{\mathrm{\textit{T}}}, \boldsymbol{x}_{\mathrm{\textit{I}}}, \boldsymbol{\pi}_{\mathrm{\textit{I}}} \right)$.

First, the SRT predicts a low-resolution $32 \times 32$ latent image $\tilde{\boldsymbol{x}}_{\mathrm{\textit{T}}}$ based on the target view $\boldsymbol{\pi}_{\mathrm{\textit{T}}}$ for computationally efficiency. The latent image from SRT is concatenated with the noisy image $\boldsymbol{y}$ and fed into the latent diffusion UNet $\boldsymbol{\mathcal{E}}_\theta$. 
In addition, we condition 
$\boldsymbol{\mathcal{E}}_\theta$ on the latent scene representation $\boldsymbol{z}$ via cross-attention layers (see \cref{fig:Pipeline}).

The generated images $\hat{\boldsymbol{\epsilon}_t}$ can be denoted as
\begin{align}
   \hat{\boldsymbol{\mathcal{E}}_t} &= \boldsymbol{\mathcal{E}}_\theta (\boldsymbol{y}, \tilde{\boldsymbol{x}}_{\mathrm{\textit{I}}}, \boldsymbol{z}, t),
\end{align}
where $t$ is the timestep.

We optimize a simplified variational lower bound, that is

\begin{align}
  \mathcal{L}_{\mathrm{VLDM}}=\mathbb{E}\left[\left\|
      \boldsymbol{\mathcal{E}}_t - \boldsymbol{\mathcal{E}}_\theta (\boldsymbol{y}, \tilde{\boldsymbol{x}}_{\mathrm{\textit{T}}}, \boldsymbol{z}, t)
  \right\|^2\right].
\end{align}

\textbf{Multi-View Attention}. As previously stated, in Zero123~\cite{liu2023zero}, multiple images are generated in sequence from a given input view based on camera parameters. This approach can introduce inconsistencies between generated views.
To address this issue, we apply modifications to the UNet in order to feed multi-view images. This way, we can predict simultaneously multiple novel views. We employ self-attention block to ensure consistency for different viewpoints.

\section{Experiments}
\label{sec:Results}

This section presents the novel view synthesis experiments in \cref{sec:Novel View Synthesis}, and the 3D generation experiments in \cref{sec:3D Generation}. We present ablation experiments in \cref{sec:Ablation} and ethical considerations in \cref{sec:Ethics}.

\paragraph{Training Data.}
For training our model for novel view synthesis, we use 800k 3D object models from Objaverse~\cite{objaverse}. For a fair comparison with other 3D diffusion baselines, we use the same training dataset.

Input condition views are chosen in a similar way as Zero123~\cite{liu2023zero}. An azimuth angle is randomly chosen from one of the eight discrete angles of the output cameras. The elevation
angle is randomly selected in the range [$-10^{\circ}, 45^{\circ}$]. 
For data quality purposes, we discard empty rendered images. This represents about one per cent of the training data. The data filtering strategy is similar to ~\cite{Kong2024EscherNetAG}. 3D objects are centered and we apply uniform scaling in the range [-1,1] so that dimensions matches. Input images to our pipeline are RGB images 256x256. 

\paragraph{Test Data.}
We use the Google Scanned Object (GSO)~\cite{GSO} as our testing dataset, and use the same 30 objects as SyncDreamer~\cite{liu2023syncdreamer}. There are 16
images per 3D object, with a fixed elevation of $30^{\circ}$ and every $22.5^{\circ}$ for azimuth. 

\paragraph{Implementation Details.}Our model is trained using the AdamW optimiser [24] with a learning rate of $10^{-4}$ and weight decay of 0.01. We reduce the learning rate to $10^{-5}$ for a total of 100k training steps. For our training batches, we use 3 input views and 3 target views randomly sampled with replacement from 12 views for each object, with a batch size of 356. We train our model for 6 days on 4 A6000 (48GB) GPUs.

\textbf{Evaluation Metrics.} For novel view synthesis, we report the PSNR, SSIM~\cite{SSIM}, and LPIPS~\cite{LPIPS}. For 3D reconstruction from single-view or few views, we use the Chamfer Distances (CD) and 3D IoU between the ground-truth and reconstructed volumes.

\subsection{Novel View Synthesis}
\label{sec:Novel View Synthesis}

\begin{figure*}
  \centering
    \includegraphics[width=1\linewidth]{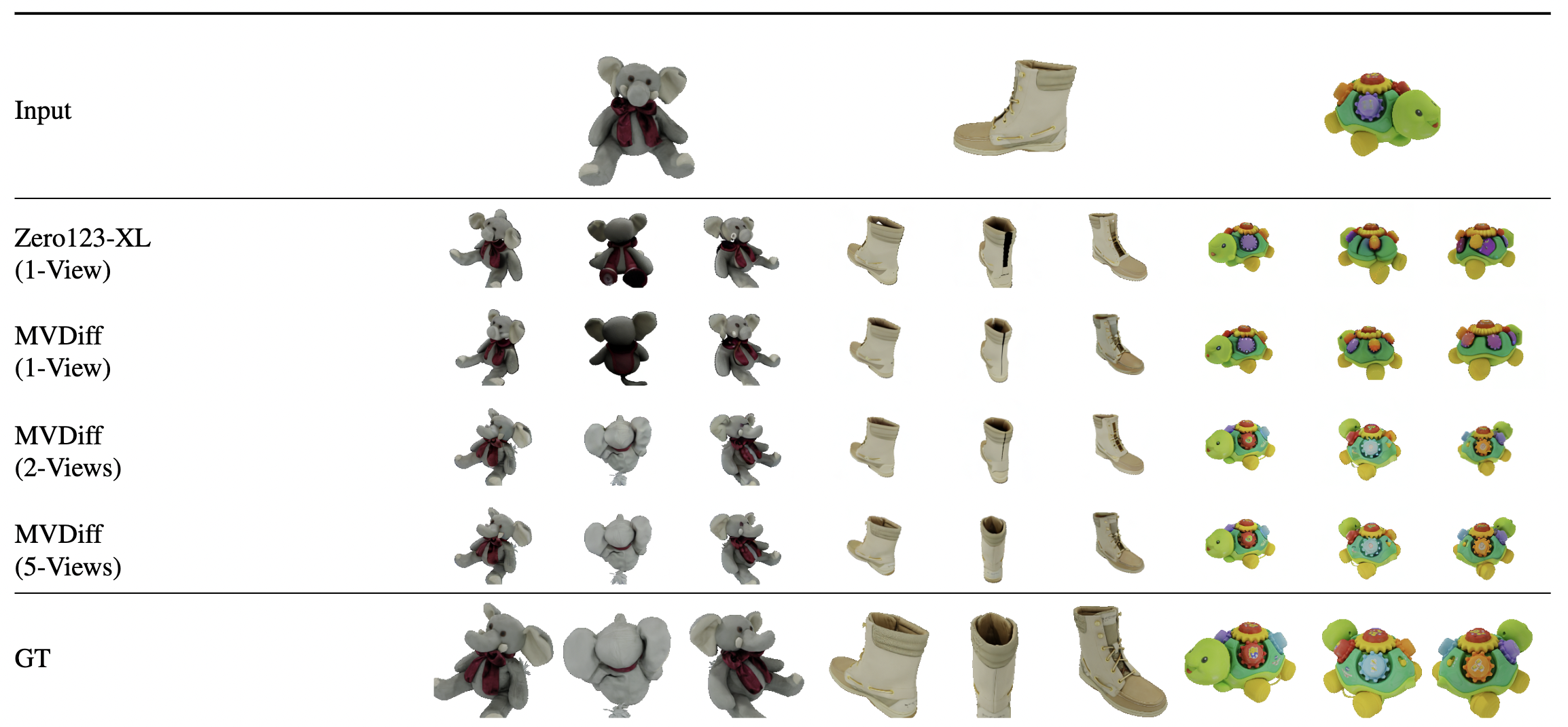}
  \caption{{\bf Zero-Shot Novel View Synthesis on GSO.} MVDiff outperforms Zero123-XL for single view generation with greater camera control and generation quality. As more views are added, MVDiff resembles the ground-truth with fine details being captured such as elephant tail and turtle shell design.}
  \label{fig:novel_view_synthesis} 
\end{figure*}

\begin{figure}[ht!]
\centering
\footnotesize
\setlength{\tabcolsep}{0.0em}
{\makecell{\small Input\\}}\\\includegraphics[trim={1cm 1cm 1cm 1cm}, clip, width=0.165\linewidth]{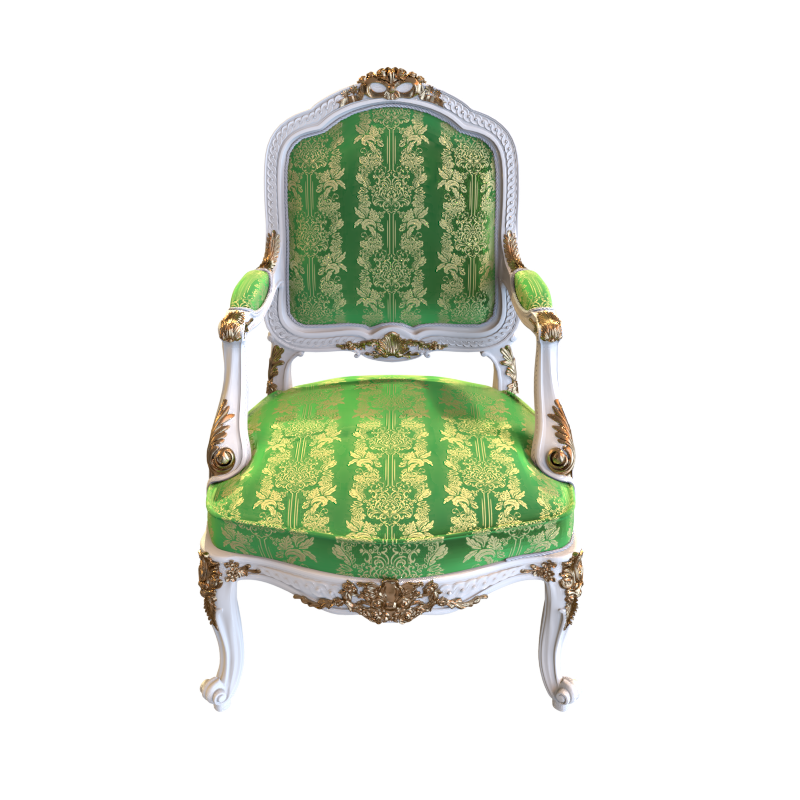}\\
\begin{tabular}{*{6}{C{0.165\linewidth}}}
{\makecell{\small $\leftarrow-----Generated-----\rightarrow$\\}}\\
{\includegraphics[trim={1cm 1cm 1cm 1cm}, clip, width=\linewidth]{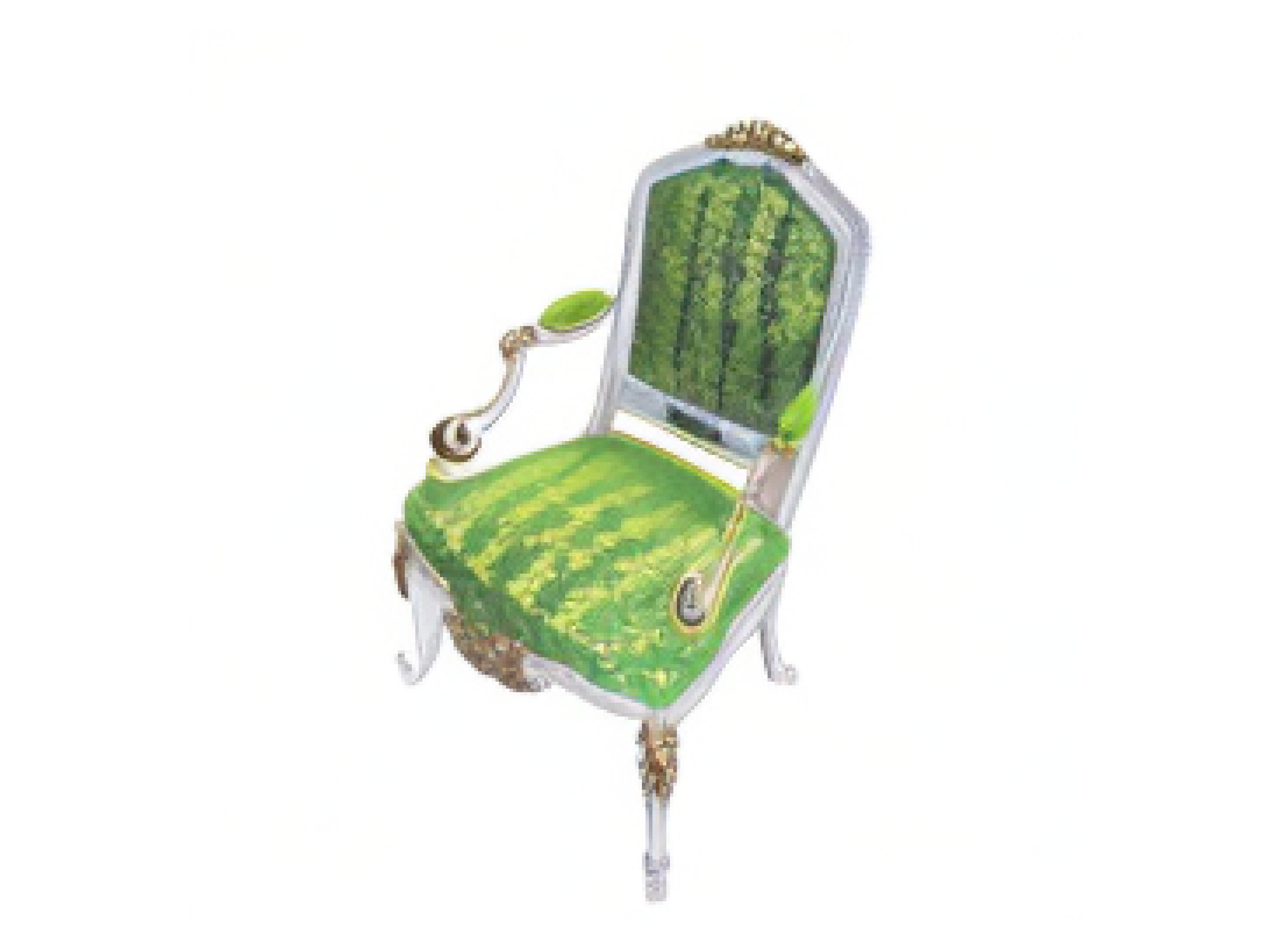}}&
\includegraphics[trim={1cm 1cm 1cm 1cm}, clip, width=\linewidth]{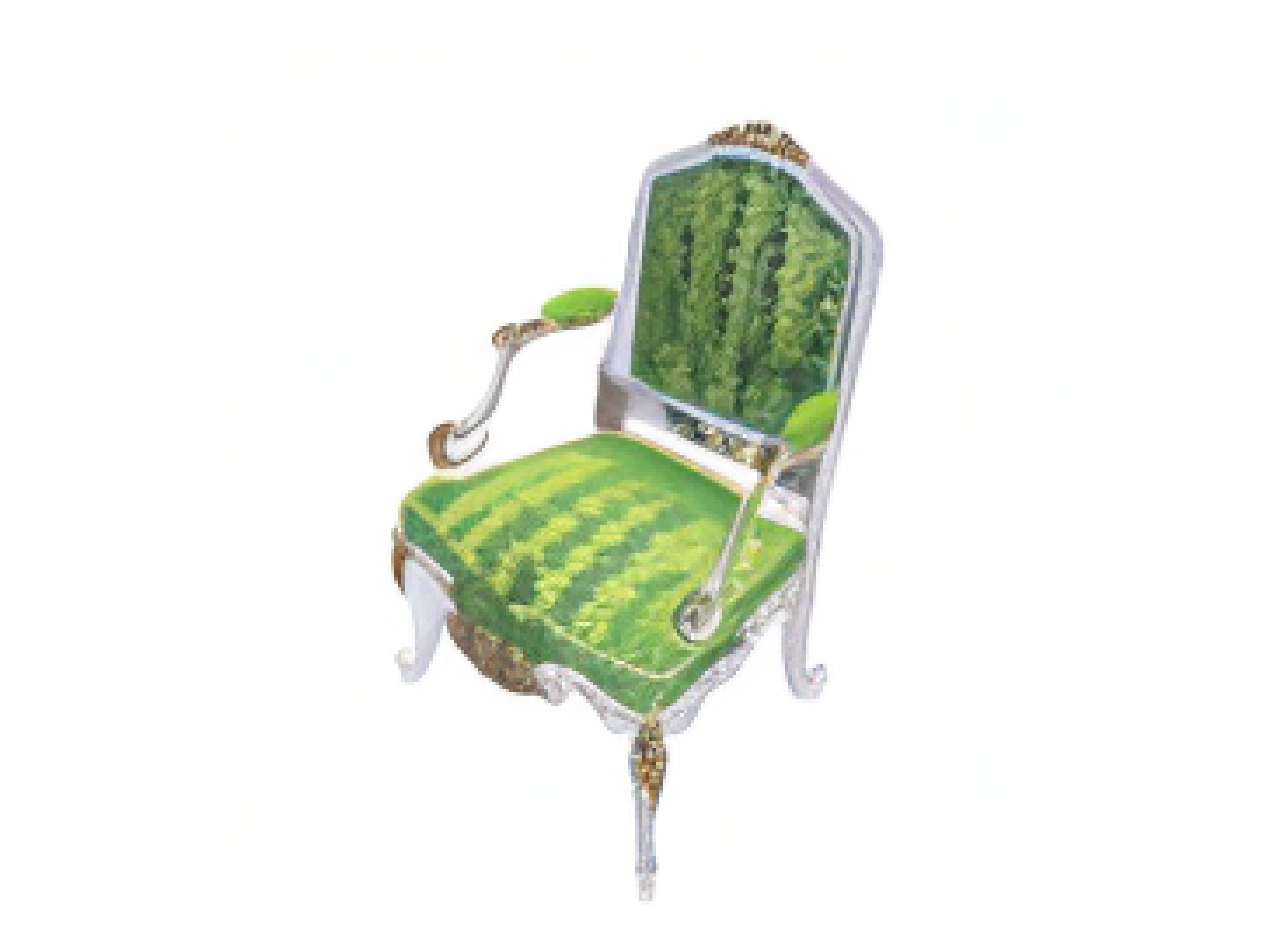}& 
\includegraphics[trim={1cm 1cm 1cm 1cm}, clip, width=\linewidth]{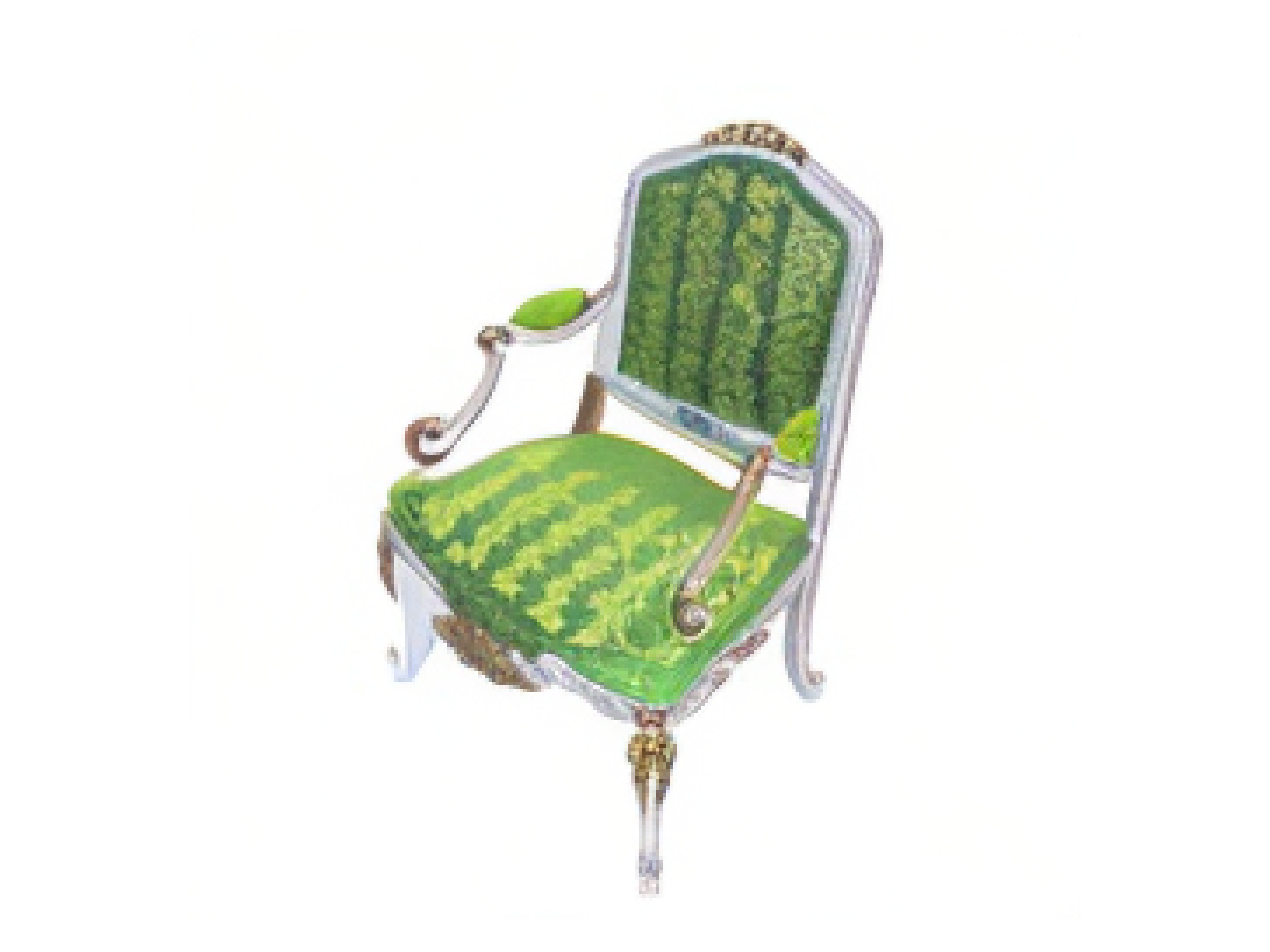}& 
\includegraphics[trim={1cm 1cm 1cm 1cm}, clip, width=\linewidth]{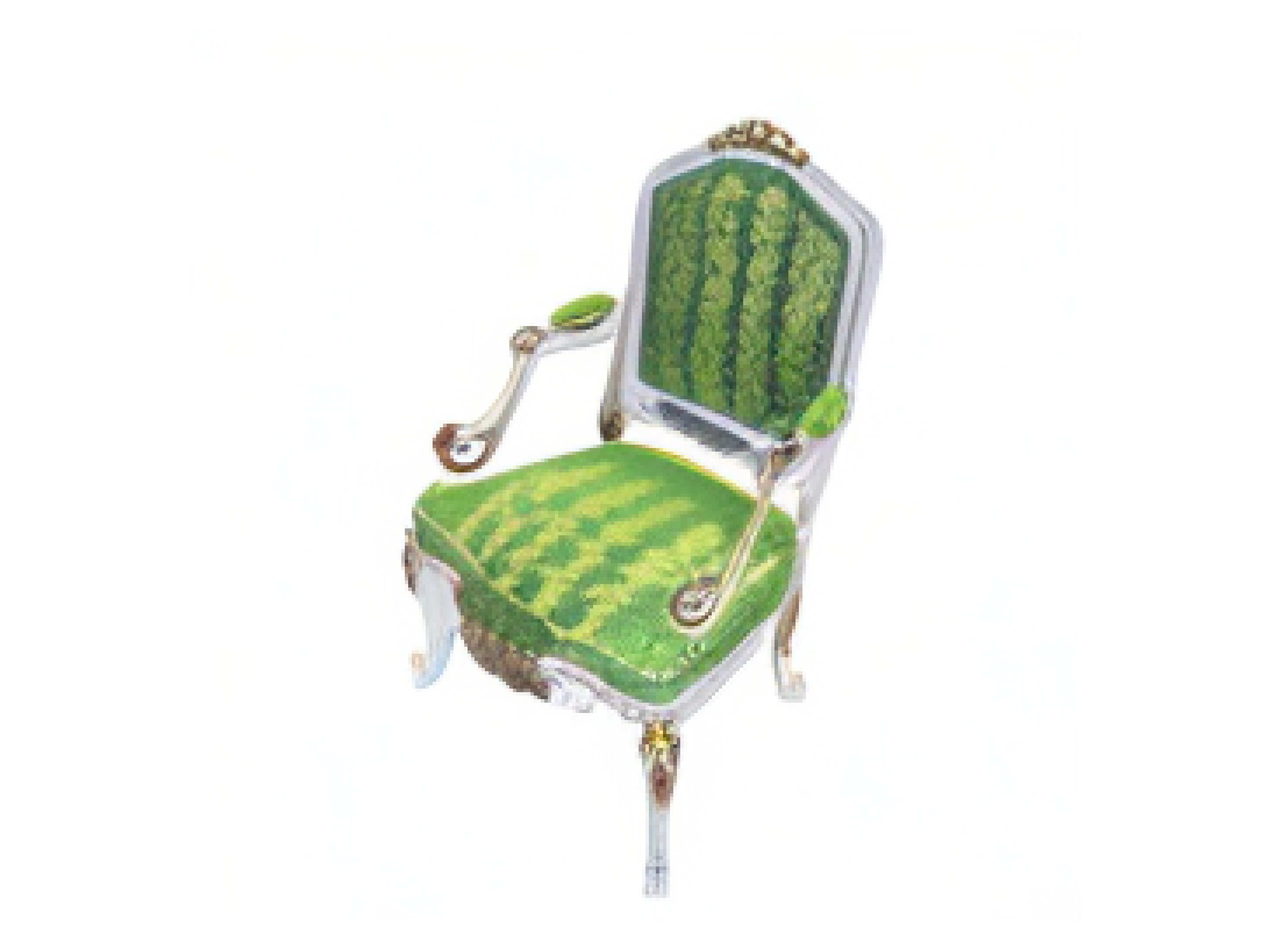} & 
{\makecell{\small  GT\\}}
{\includegraphics[trim={1cm 1cm 1cm 1cm}, clip, width=0.9\linewidth]{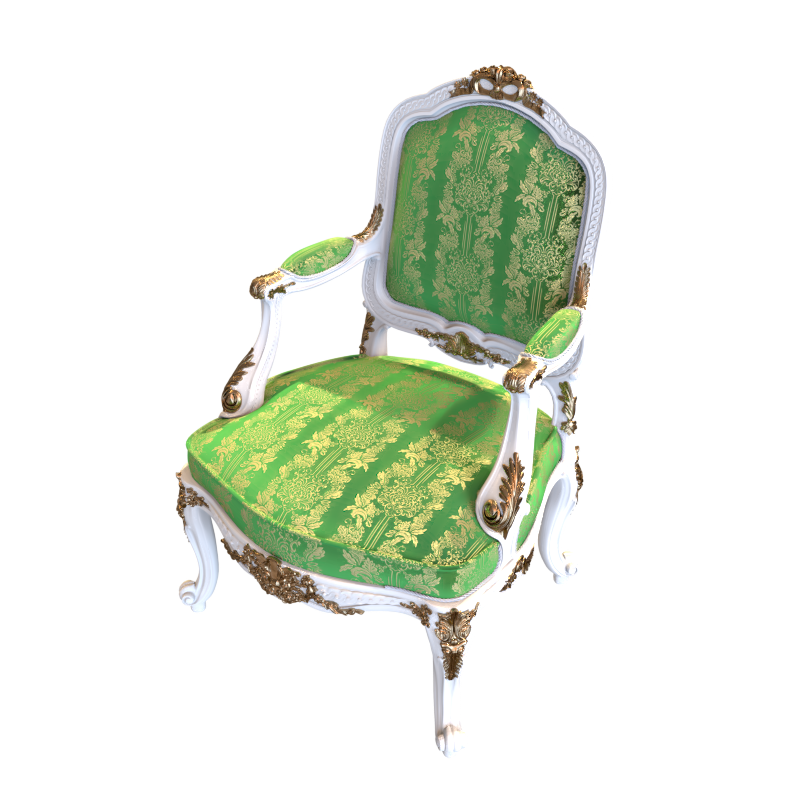}}\\
{\includegraphics[trim={1cm 1cm 1cm 1cm}, clip, width=\linewidth]{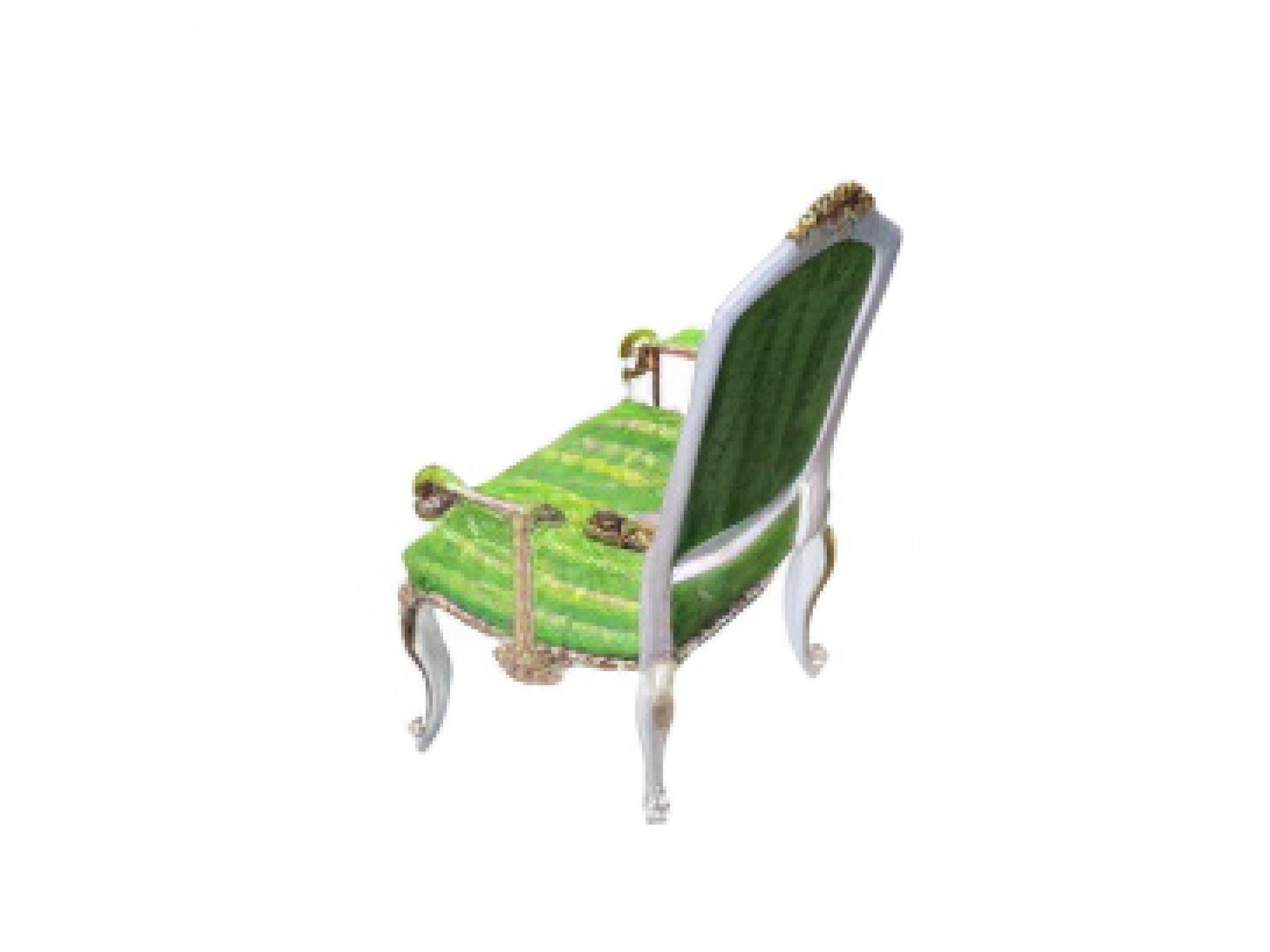}}&
\includegraphics[trim={1cm 1cm 1cm 1cm}, clip, width=\linewidth]{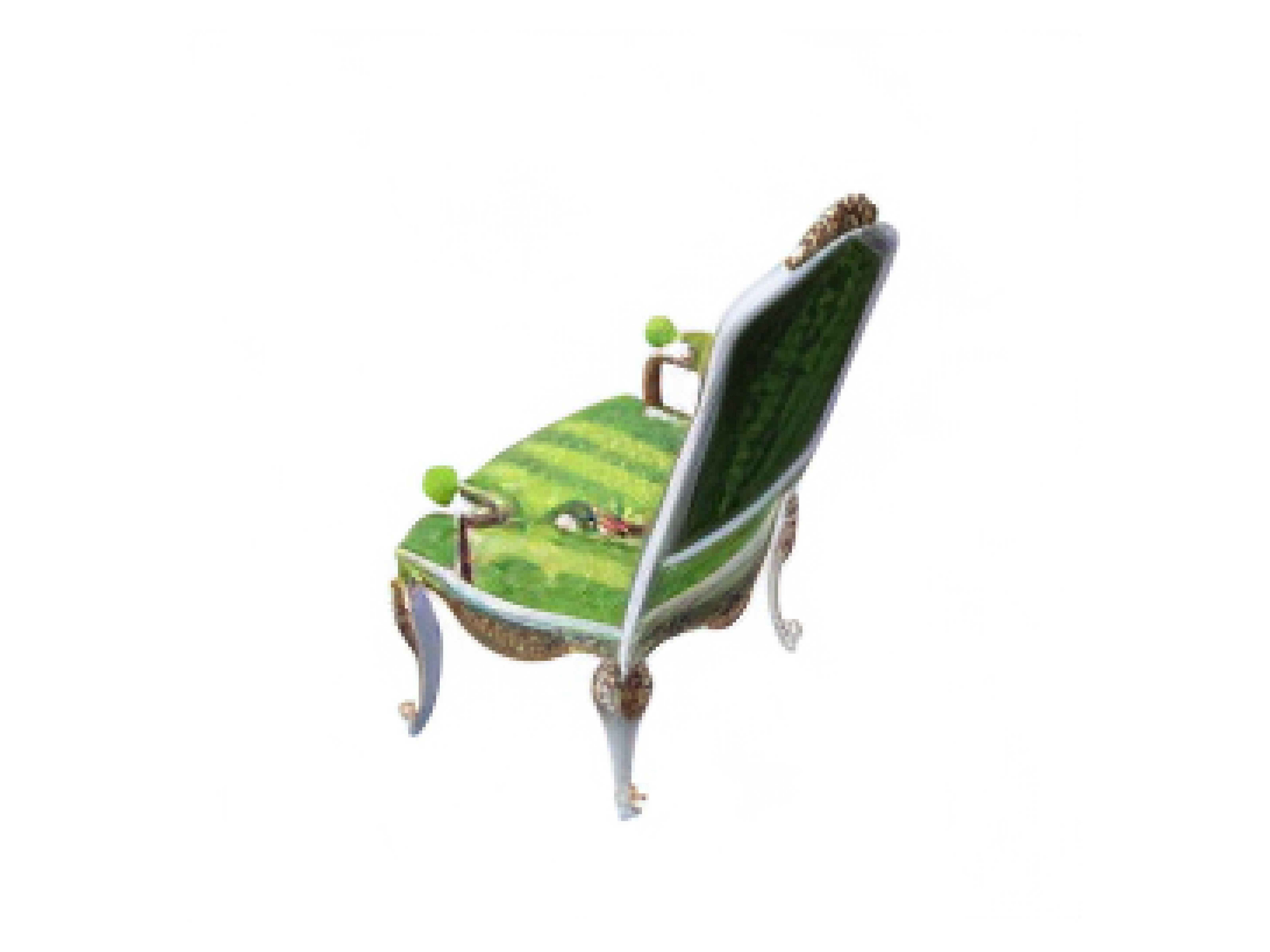}& 
\includegraphics[trim={1cm 1cm 1cm 1cm}, clip, width=\linewidth]{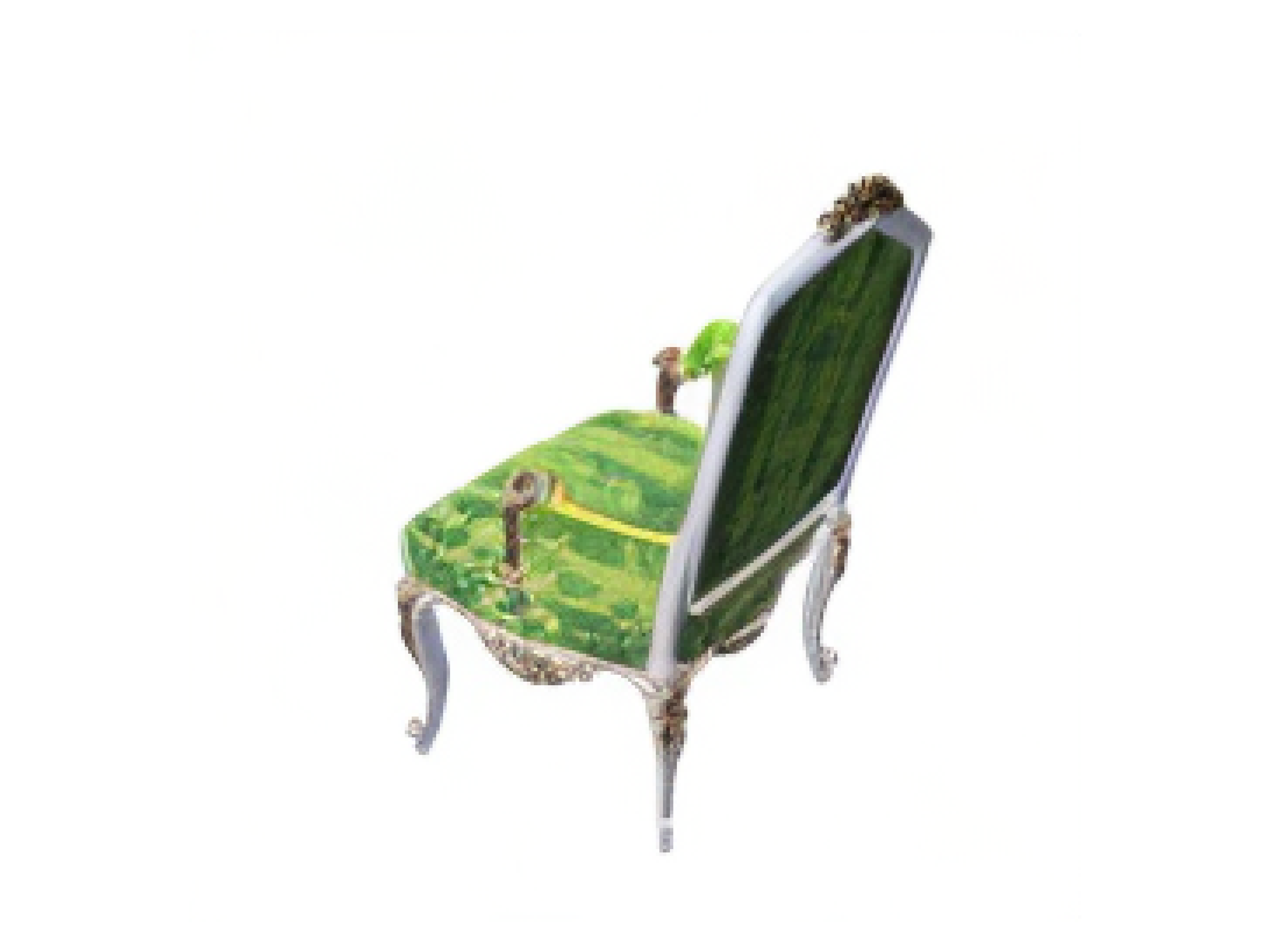}& 
\includegraphics[trim={1cm 1cm 1cm 1cm}, clip, width=\linewidth]{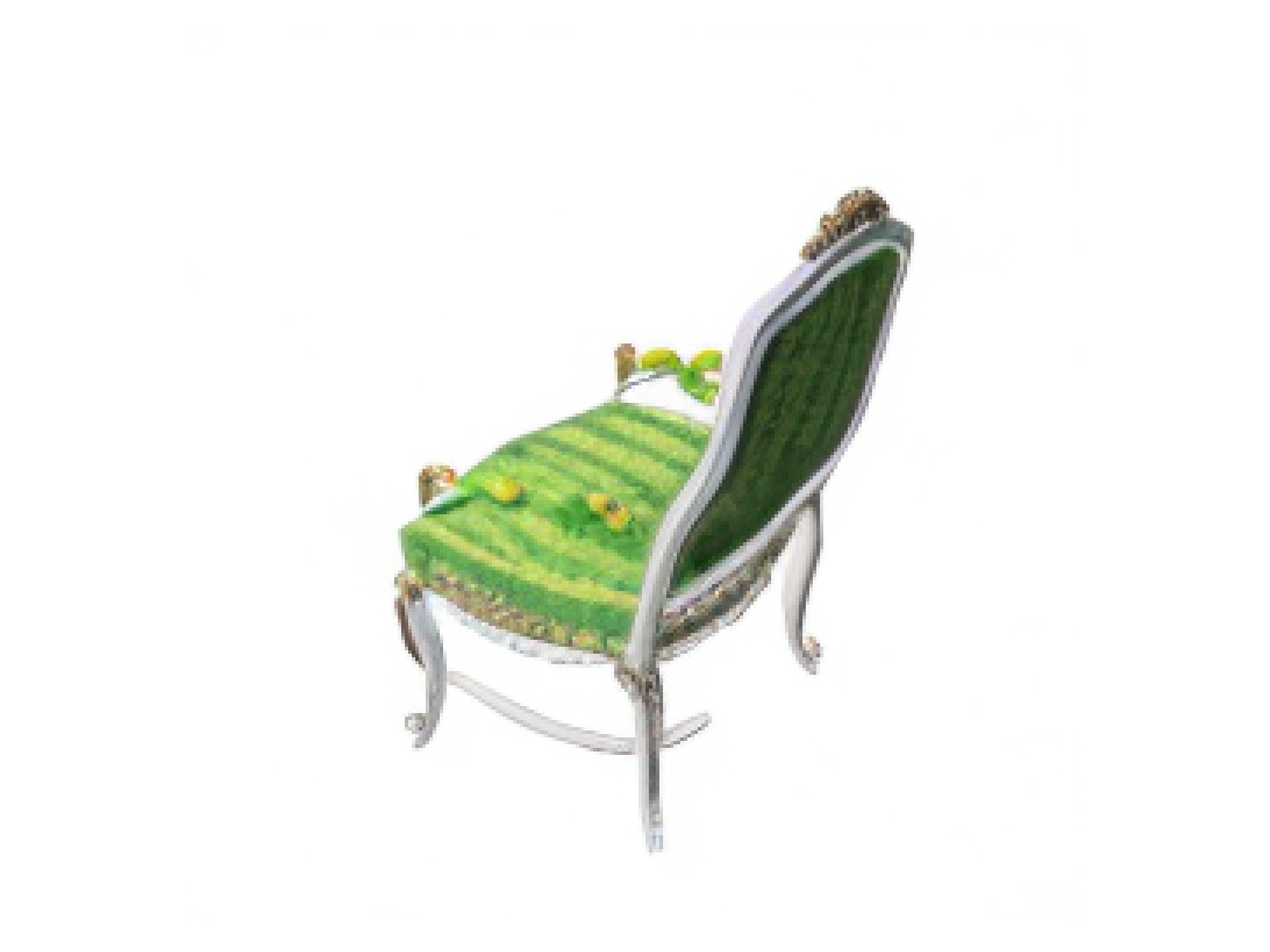}& 
{\includegraphics[trim={1cm 1cm 1cm 1cm}, clip, width=\linewidth]{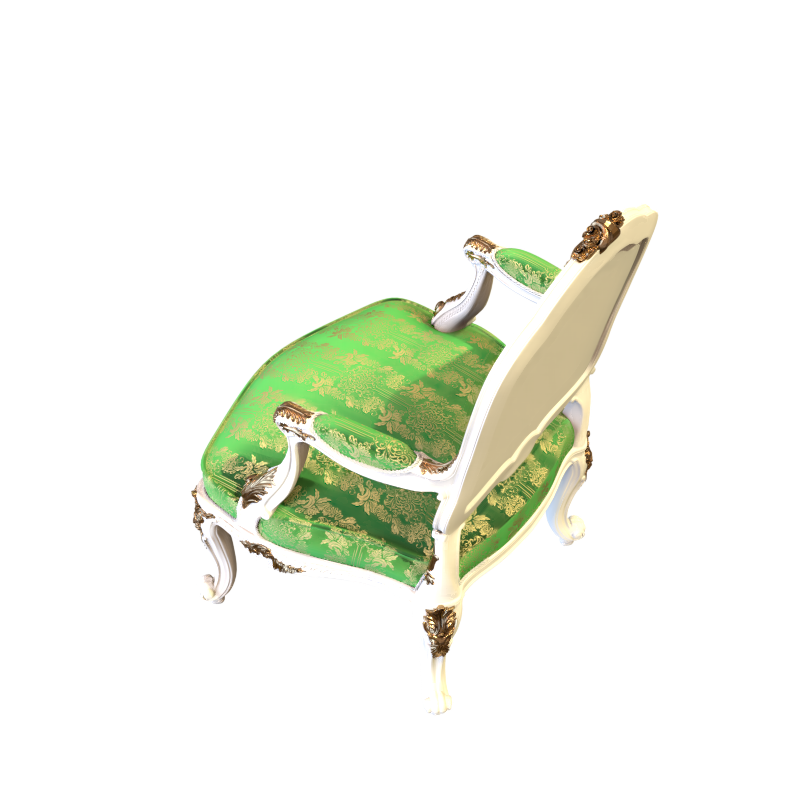}}\\
{\includegraphics[trim={1cm 1cm 1cm 1cm}, clip, width=\linewidth]{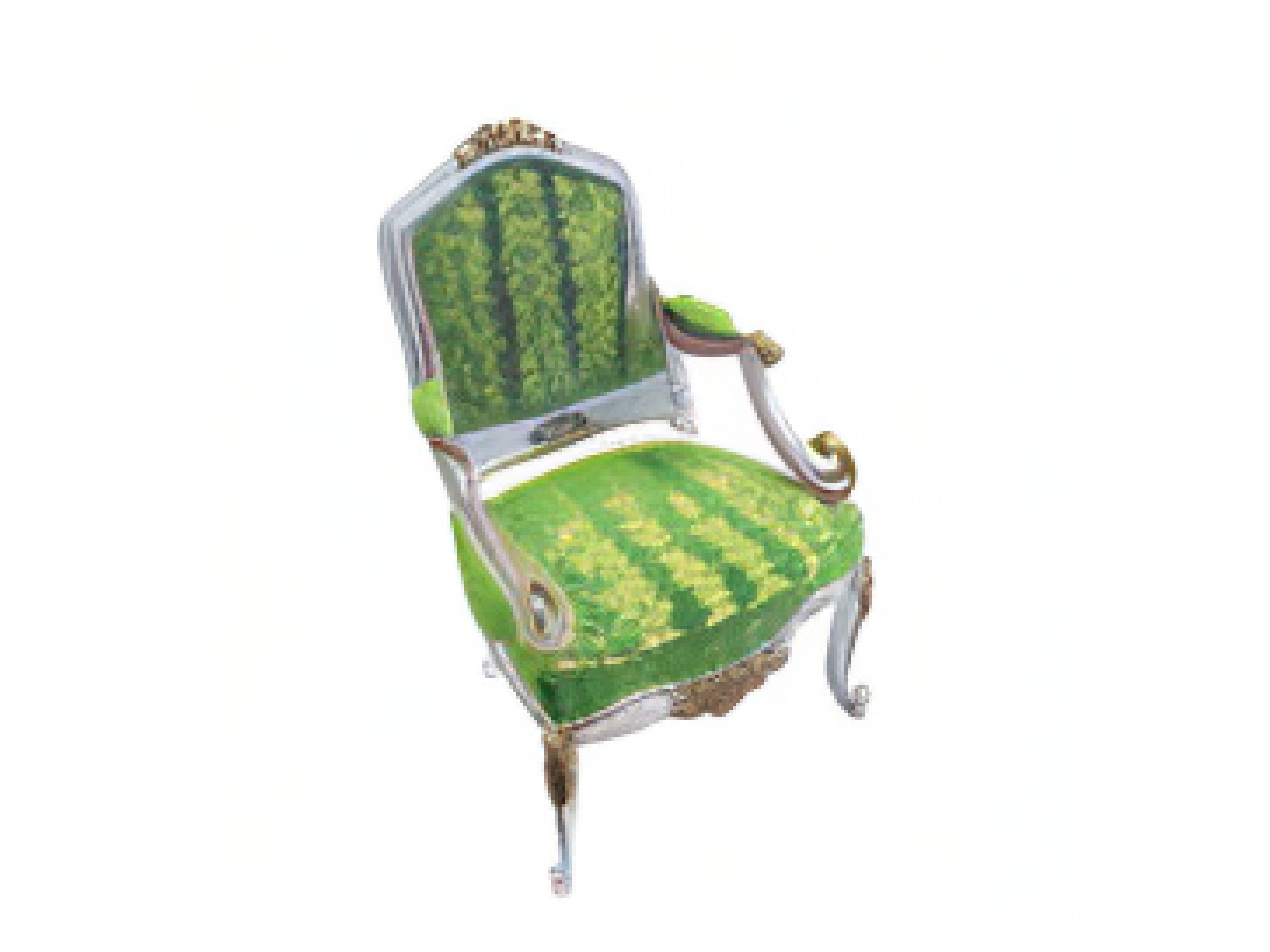}}&
\includegraphics[trim={1cm 1cm 1cm 1cm}, clip, width=\linewidth]{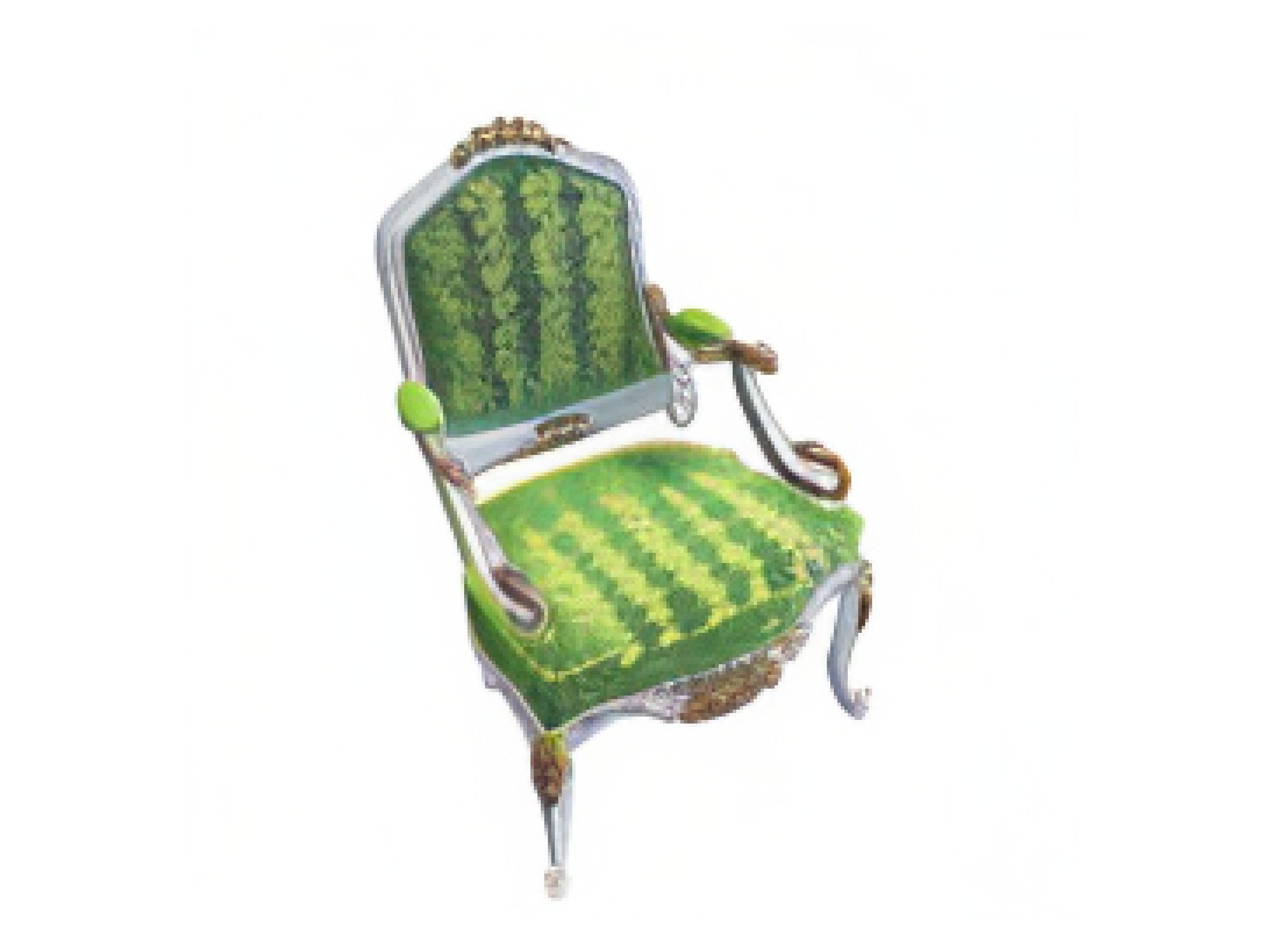} & 
\includegraphics[trim={1cm 1cm 1cm 1cm}, clip, width=\linewidth]{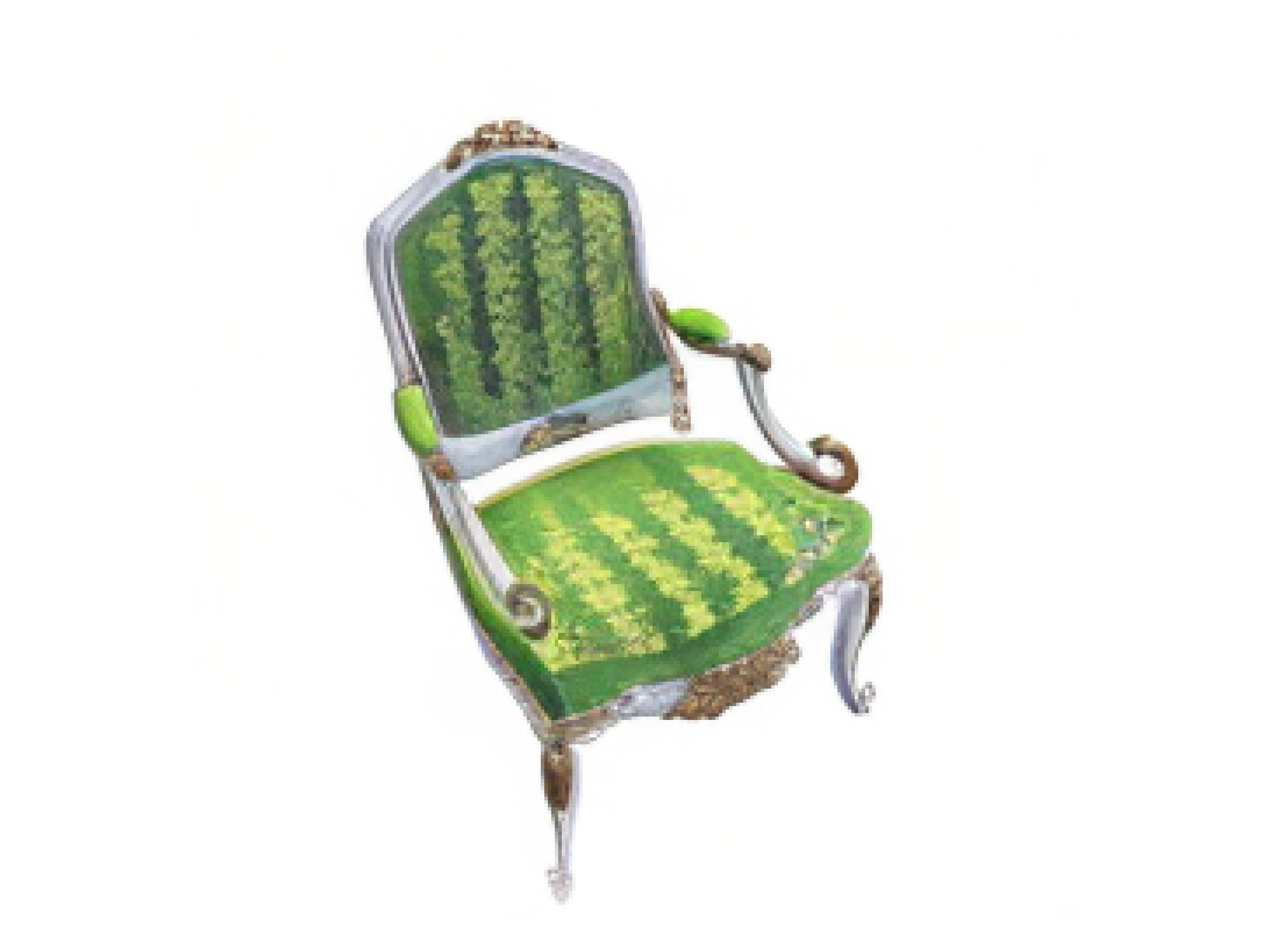} & 
\includegraphics[trim={1cm 1cm 1cm 1cm}, clip, width=\linewidth]{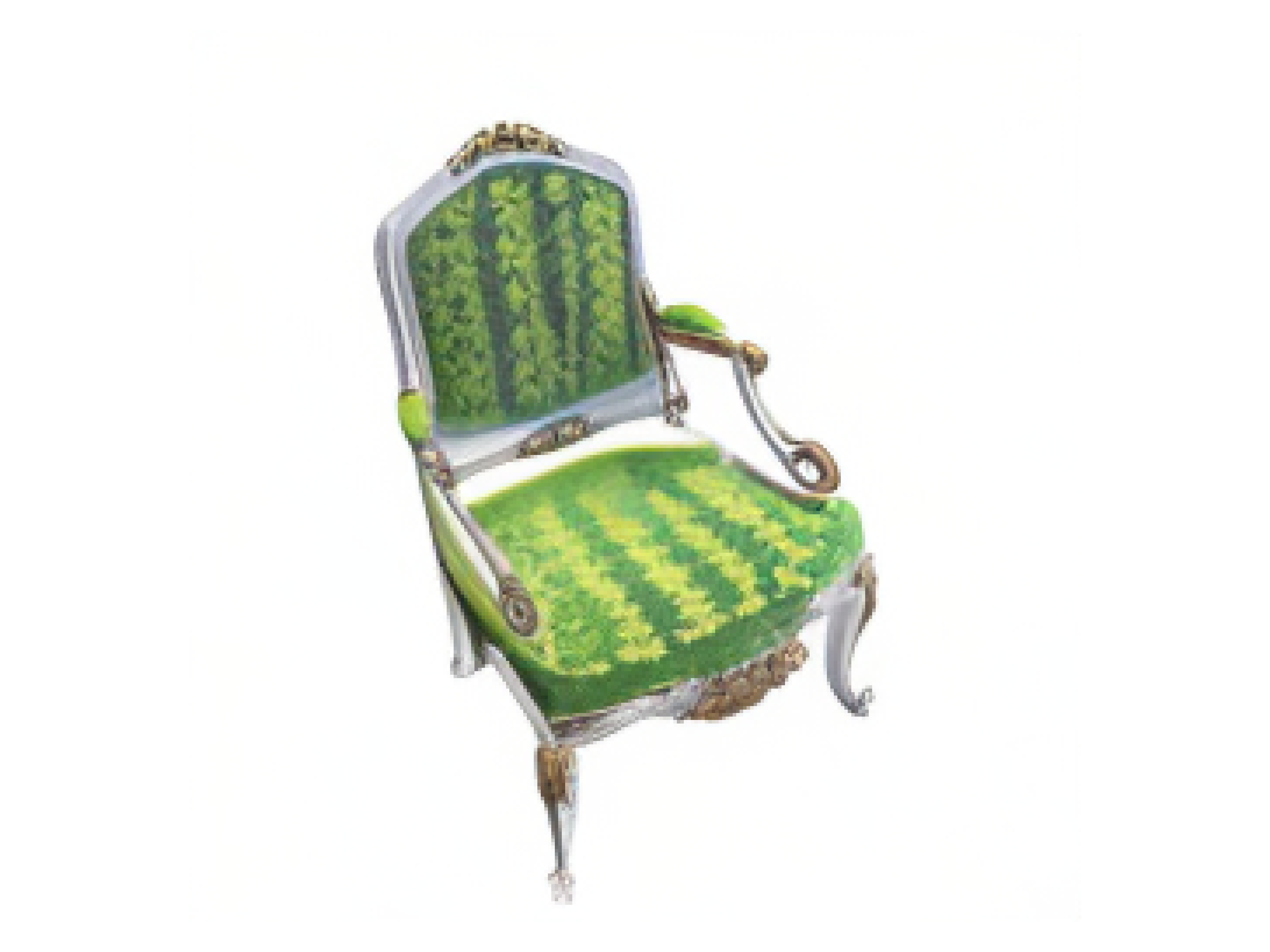} &   
{\includegraphics[trim={1cm 1cm 1cm 1cm}, clip, width=\linewidth]{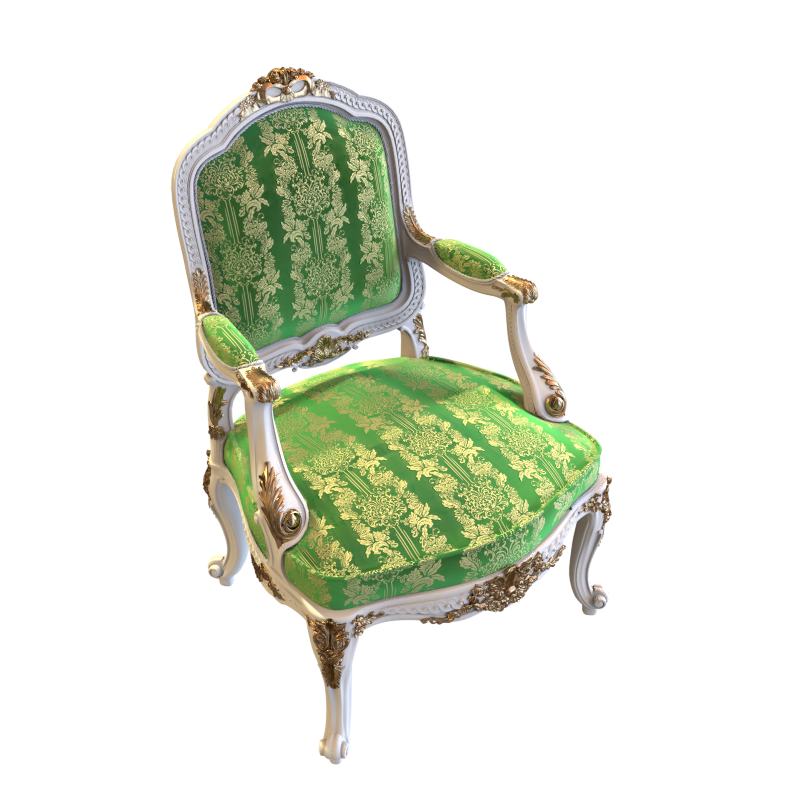}}
\end{tabular}
\caption{{\bf Diversity of Novel View Diffusion with MVDiff on NeRF-Synthetic Dataset.} We show nearby views (\textit{top and bottom row}) displaying good consistency, while more distant views (\textit{middle}) are more diverse but still realistic.}
\label{figure:Diversity}
\end{figure}

We show in \cref{table:novel_view_synthesis} the performance of MVDiff compared to baselines for novel view synthesis on an unseen dataset~\cite{GSO}. Qualitative results are shown in \cref{fig:novel_view_synthesis}. Our model surpasses baseline Zero-123XL by a margin and benefits from additional views. Given the probabilistic nature of the model, it is able to generate diverse and realistic shapes given a single view (see \cref{figure:Diversity}).

\begin{table}[ht]
   \centering
   \scriptsize
   \setlength{\tabcolsep}{0.2em}
   \resizebox{0.45\textwidth}{!}{ 
       \begin{tabular}{lccccccccccc}
       \toprule
       &  \multirow{2}[2]{*}{\makecell[c]{Training \\ Sample}} & \multirow{2}[2]{*}{\makecell{\# Ref.\\ Views}} & \multicolumn{4}{c}{{\bf GSO}}  & \multicolumn{4}{c}{{\bf  NeRF Synthetic}} \\
       \cmidrule(lr){4-7}      \cmidrule(lr){8-11}
          & &  &PSNR$\uparrow$ & SSIM$\uparrow$ & LPIPS$\downarrow$& Runtime$\downarrow$   &PSNR$\uparrow$ & SSIM$\uparrow$ & LPIPS$\downarrow$& Runtime$\downarrow$ \\
        \midrule 
        Zero123\cite{liu2023zero}  &  800K  &1  &18.51 &0.856 &0.127  & 7s &12.13 &0.601 &0.421& 7s\\
        Zero123-XL\cite{liu2023zero} & 10M  &1 &18.93 &0.856 &0.124 & 8s &12.61 &0.620 &0.381 & 8s\\
        EscherNet\cite{Kong2024EscherNetAG} & 800k &1 &20.24 &0.884 &0.095 & 8s & 13.36 & 0.659 & 0.291 & 9s \\
        MVDiff & 800k  &1  &20.24 &0.883 &0.094 & 9s &12.66 &0.638 &0.342 & 9s\\
        MVDiff   & 800k  &2  &22.92 &0.91 &0.063 &
        9s &13.42 &0.685 &0.321 & 10s\\
        MVDiff    & 800k  &3  &24.11 &0.921 &0.049 &10s &13.58 &0.741 &0.301 & 11s\\
        MVDiff   & 800k &5  &25.24 &0.931 &0.040 &11s &14.55 &0.833 &0.288 & 12s\\
        MVDiff  & 800k &10   &25.94 &0.937 &0.034 & 12s &14.51 &0.657 &0.215 & 13s\\
         \bottomrule
       \end{tabular}
    }
   \caption{{\bf Novel view synthesis performance on GSO and NeRF Synthetic datasets.} MVDiff outperforms Zero-123 as well as Zero-123XL with significantly less training data. MVDiff shows improved performance with the addition of more reference views.}
   \label{table:novel_view_synthesis}
   \vspace{-0.4cm}
\end{table}

\subsection{3D Generation}
\label{sec:3D Generation}

We showed in \cref{sec:Novel View Synthesis} that our model can generate multiple consistent novel views. In this section, we perform single and few-images 3D generation on the GSO dataset. 
We generate 16 views with azimuths uniformly distributed in the range \ang{0} to \ang{360}. For a fixed elevation angle of \ang{30}, SyncDreamer may fail to recover the shape of 3D objects at the top and bottom since the camera angle does not cover those regions. Therefore, we also use different elevation angles from \ang{-10} to \ang{40}. Then, we adopt NeuS~\cite{NeuS} for 3D reconstruction. The foreground masks of the generated images are initially predicted using CarveKit. It takes around 3 minutes to reconstruct a textured mesh. 

\begin{figure*}
  \centering
    \includegraphics[width=1\linewidth]{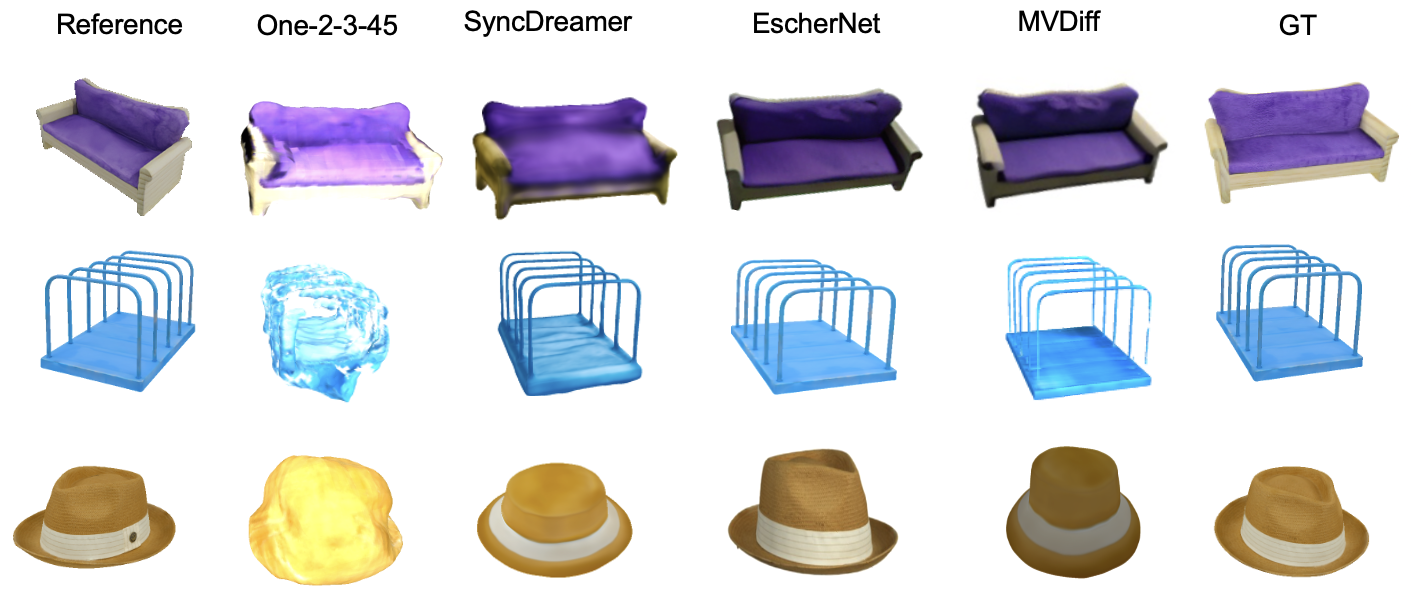}
  \caption{{\bf 3D reconstruction from single-view on GSO dataset.} MVDiff produces consistent novel views and improves the 3D geometry compared to baselines. One-2-3-45 and SyncDreamer tend to generate overly-smoothed and incomplete 3D objects, in particular the sofa. EscherNet recovers more of the finer details, as for the hat.}
  \label{fig:3d} 
\end{figure*}

We compare our 3D recontructions with SoTA 3D generation models, including One-2-3-45~\cite{liu2023one2345} for decoding an SDF using multiple views predicted from Zero123, and SyncDreamer~\cite{liu2023syncdreamer} for fitting an SDF using NeuS~\cite{NeuS} from 16 consistent fixed generated views. Given two or more reference views, MVDiff outperforms all other baselines (see \cref{table:3D_reconstruction}). MVDiff generates meshes that are visually consistent and resembles the ground-truth (see \cref{fig:3d}).

\begin{table}[ht]
   \centering
   \scriptsize
 \renewcommand{\arraystretch}{0.8}
   \setlength{\tabcolsep}{0.7em}
   \begin{tabular}{lccc}
   \toprule
     &  \makecell{\# Input Views}  & \makecell{Chamfer Dist. $\downarrow$} & \makecell{Volume  IoU $\uparrow$} \\
    \midrule 
    Point-E\cite{nichol2022pointe} &1 &0.0561 &0.2034\\
    Shape-E\cite{jun2023shape} &1 &0.0681 &0.2467 \\
    One2345\cite{liu2023one2345} &1 &0.0759 &0.2969 \\
    LGM\cite{LGM} & 1 &0.0524 &0.3851 \\
    SyncDreamer\cite{liu2023syncdreamer} & 1&0.0493 &0.4581 \\
    EscherNet\cite{Kong2024EscherNetAG} & 1 &0.0314 &0.5974 \\
    \midrule
    MVDiff & 1 &0.0411 &0.4357 \\
    MVDiff & 2 &0.0341 &0.5562 \\
    MVDiff & 3 &0.0264 &0.5894 \\
    MVDiff & 5 &0.0252 &0.6635 \\
    MVDiff & 10 &0.0254 &0.6721 \\
     \bottomrule
   \end{tabular}
   \caption{{\bf 3D reconstruction performance on GSO dataset.} MVDiff surpasses most single-view to 3D benchmarks. Note that the performance improves as the number of input views increases. }
    \label{table:3D_reconstruction}
 \end{table}

\subsection{Ablation Study}
\label{sec:Ablation}

\paragraph{Multi-View Consistency.}
The generated images may not always plausible and we need to generate multiple instances with different seeds and select a desirable instance for 3D reconstruction based on higher overall PSNR, SSIM and LPIPS for the view generated. Experiments show that we need 5 generations to obtain optimal reconstruction. 

\paragraph{Effect of Epipolar and Multi-View Attention.} We evaluate the benefits of epipolar attention and multi-view attention on novel view synthesis performing ablation experiments on those components. In particular, we observe a significant drop in performance metrics when removing epipolar attention suggesting that the model is effectively able to implicitely learn 3D object geometry by enforcing geometrical guidance (see \cref{table:ablation}).

\begin{table}\small
    \centering
    \begin{tabular}{@{}lcccc@{}}
   \toprule
    & PSNR$\uparrow$ & SSIM$\uparrow$ & LPIPS$\downarrow$ \\
   \midrule
   MVDiff & 20.24 & 0.884 & 0.095 \\
   w/o epipolar att. & 19.14 & 0.864 & 0.118 \\
   w/o multi-view att. & 19.92 & 0.871 & 0.113 \\
   \bottomrule
\end{tabular}
\caption{{\bf Effect of Self-Attention Mechanisms}. We report PSNR, SSIM~\cite{SSIM}, and LPIPS~\cite{LPIPS} for novel view synthesis from single view on GSO dataset. Results show that epipolar attention and multi-view attention lead to superior performance.}
\label{table:ablation}
\end{table}

\paragraph{Weight Initialisation.} An alternative to initialising weights trained from Zero123 on view-dependent objects~\cite{objaverseXL} is to use weights from Stable Diffusion~\cite{rombach2021highresolution}. 
We compare the performance of our model initializing weights from Stable Diffusion v2~\cite{rombach2021highresolution} with a drop in performance of -2.58 PSNR compared to Zero123~\cite{liu2023zero} weight initialisation. This shows that initializing from Stable Diffusion v2 leads to poorer performance on the novel view task and worse generalisability.

\subsection{Risks and Ethical Considerations}
\label{sec:Ethics}
There are several promising applications of synthetic data, notably in medicine. Synthetic data could make significant improvement in surgery planning and tailored patient diagnosis leveraging 3D information and its assets of quantitative parameters. Nevertheless, there are ethical considerations associated with the use of synthetic data in medicine. We should ensure the synthetic data is anonymised such that no particular features of the synthetic meshes could link back to a specific patient. In that light, there are transformations that can be applied to the meshes. We should also make sure that the synthetic data is not used in a way it could harm or be detrimental. Further validation on different cohorts of people is required before using these synthetic data in clinical settings. 

Despite important ethical considerations we shed light on, we believe these 3D representations of organs could be of great use, on hand for research purposes to run large-scale statistical analysis on different cohorts and highlight associations with patient metadata. These cost effective synthetic data could be beneficial to improve the visualisations of bones and organs and be deployed widely.

\subsection{Limitations}

A limitation of this work lies in its computational time and resource requirements. Despite advances in sampling approaches, our model still requires more than 50 steps to generate high-quality images. This is a limit of all diffusion based generation models. Moreover, the reconstructed meshes may not always be plausible. To increase the quality, we may need to use a larger object dataset like Objaverse-XL\cite{objaverseXL} and manually curate the dataset to filter out uncommon shapes such as point clouds, textureless 3D models and more complex scene representation.

\section{Conclusion}
\label{sec:Conclusion}

In our work, we aimed to address the problem of inconsistencies in multi-view synthesis from single view. We specifically apply epipolar attention mechanisms as well as multi-view attention to aggregate features from multiple views. We propose a simple and flexible framework capable of generating high-quality multi-view images conditioned on an arbitrary number of images.

\subsection{Future Work}
\label{sec:Future Work}

\paragraph{Combining with graphics.}
In this study, we show that we can generate view consistent 3D objects by learning geometrical correspondences between views during training. We modified the latent diffusion U-Net model to feed multi view in order to generate consistent multi view for 3D reconstruction. Future work can explore utilising knowledge about lighting, and texture to generate more diverse range of 3D shapes with varying lighting and texture.

\subsection*{Acknowledgements}
\label{sec:Acknowledgements}
E.B is supported by the Centre for Doctoral Training in Sustainable Approaches to Biomedical Science: Responsible and Reproducible Research (SABS: R3), University of Oxford (EP/S024093/1). P.B. is supported by the UKRI CDT in AI for Healthcare http://ai4health.io (Grant No. P/S023283/1). We were inspired by the tables design of Eschernet\cite{Kong2024EscherNetAG} and we thank the authors for their great work.
{
    \small
    \bibliographystyle{ieeenat_fullname}
    \bibliography{main}
}
\end{document}